\documentclass{article}

\usepackage[final]{nips_2016}

\usepackage[utf8]{inputenc} % allow utf-8 input
\usepackage[T1]{fontenc}    % use 8-bit T1 fonts
\usepackage{hyperref}       % hyperlinks
\usepackage{url}            % simple URL typesetting
\usepackage{booktabs}       % professional-quality tables
\usepackage{amsfonts}       % blackboard math symbols
\usepackage{nicefrac}       % compact symbols for 1/2, etc.
\usepackage{microtype}      % microtypography
\usepackage{bm}
\usepackage{amsmath}
\usepackage{graphicx}
\usepackage{wrapfig}
\usepackage{multirow}

\title{Predictive Coding for Dynamic Vision: \\Development of Functional Hierarchy in a Multiple Spatio-Temporal Scales RNN Model}
\author{
  Minkyu Choi \qquad\quad  Jun Tani\\
  Department of Electrical Engineering\\
  Korea Advanced Institute of Science and Technology\\
  \texttt{ \{minkyu.choi8904, tani1216jp\}@gmail.com} \\
  The correspondence should be sent to Jun Tani.
}

\begin{document}
% \nipsfinalcopy is no longer used

\maketitle

\begin{abstract}
 The current paper presents a novel recurrent neural network model, predictive multiple spatio-temporal scales RNN (P-MSTRNN), which can generate as well as recognize dynamic visual patterns in a predictive coding framework. The model is characterized by multiple spatio-temporal scales imposed on neural unit dynamics through which an adequate spatio-temporal hierarchy develops via learning from exemplars. The model was evaluated by conducting an experiment of learning a set of whole body human movement patterns, which was generated by following a hierarchically defined movement syntax. The analysis of the trained model clarifies what types of spatio-temporal hierarchy develops in dynamic neural activity as well as how robust generation and recognition of movement patterns can be achieved by using the error minimization principle.
\end{abstract}

\section{Introduction}
Predictive coding is a brain plausible principle to account for how diversity of perceptual input sequences can be predicted or generated for different intentions in the top-down pathway as well as how the corresponding intention can be inferred or recognized for a particular observation of perceptual sequence in the bottom-up pathway by using the prediction error minimization principle. Within this predictive coding framework, it has been largely assumed that such prediction/generation and inference/recognition can be conducted through multiple levels across different cortical areas whereas necessary functional hierarchy can be developed via accumulated learning [1, 2, 3, 4]. The current study examines how a spatio-temporal hierarchy that is adequate for robust generation and recognition of compositional dynamic visual patterns in the pixel level can be developed by proposing a novel deterministic predictive coding type deep recurrent neural network model.

The proposed model, referred to as the predictive Multiple Spatio-Temporal scales RNN (P-MSTRNN) is a combination of prior-proposed models of the multiple timescales RNN (MTRNN) [4] and the deconvolutional neural network model [15] where both spatial and temporal multiple scales properties are used as macroscopic constraints to develop an effective functional hierarchy. There have been prior studies on predicting and generating dynamic visual datasets [6, 7, 12, 13]. In particular, similar to the current paper, Lotter et al. in 2016 [13] also adopted predictive coding in their work.
However, P-PMSTRNN differs from the previously proposed generative models for dynamic vision processing, because in the current model not only prediction in the top-down pathway but also inference in the bottom-up pathway can be performed.
Also, P-MSTRNN is advantageous because it can deal with temporal hierarchy in addition to the spatial hierarchy by using the multiple temporal scales properties in the model. 
The current study is assumed to be the first attempt to achieve both generation/prediction and recognition/inference of dynamic visual images within one neural network model. 

The model was tested with using a video dataset consisting of multiple types of human movement patterns generated by multiple subjects. After the model was trained for multiple movement patterns, we examined how those multiple patterns are regenerated by using a functional hierarchy self-organized during the learning. We also scrutinized how concatenation of the learned movement patterns is developed further via additional learning. Furthermore, we examined how test movement patterns are recognized by inferring the corresponding intention. This recognition test was conducted by following a test framework, referred to as imitative synchronization [11], where P-MSTRNN attempts to imitate perceived test movement patterns. We closely analyzed how the model network recognizes possible shifts in test movement patterns by inferring the corresponding shifts of intentions using the on-line error regression scheme. 
The video dataset used in training and testing consists of multiple cyclic human movement patterns performed by multiple subjects. A set of simulation experiments examined the spatio-temporal hierarchy as it develops during learning, and revealed that robust recognition of test patterns depends on learning spatial-temporal interdependencies with exemplar patterns exhibiting high degrees of variance.

\section{Model}
\label{gen_inst}
The newly proposed model, the predictive multiple spatio-temporal scales RNN (P-MSTRNN) is a hierarchical neural network model that utilizes spatio-temporal features for generating and recognizing dynamic visual images. The P-MSTRNN model is based on the multiple spatio-temoral scales RNN (MSTRNN) [8]. The original MSTRNN is a classification model for dynamic visual images without prediction mechanism. P-MSTRNN differs from MSTRNN because it learns, generates, and recognizes patterns using the principle of prediction error minimization in the predictive coding framework.
\subsection{Architecture}
P-MSTRNN consists of a series of context layers and one input/output layer in the bottom, as shown in Figure 1.
Each context layer predicts its own neural state at each next time step by receiving the top-down and bottom-up signals from the neighboring layers. The neural state of each context layer represents its top-down intention. The first context layer predicts the next step visual inputs by receiving both the current visual input and the top-down signal from the second layer. In the upward direction, the prediction error signal back-propagates inversely in the same pathway for the purpose of updating the connectivity weights as well as the intention at each context layer in the learning and recognition process. 
%%%%%%%%%%%%%%%%%%%%%%%%%%%%%%%%%%%%% FIGURE 1 %%%%%%%%%%%%%%%%%%%%%%%%%%%%%%%%%%%%%%
\begin{figure}[h]
  \centering 
  \includegraphics[scale=0.3]{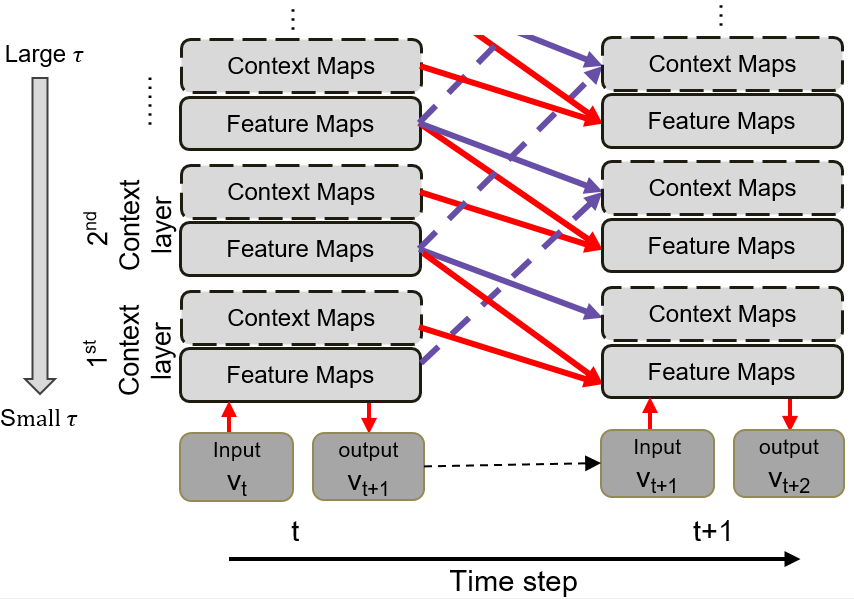}%[scale=0.3]
  \caption{Structure of the proposed network model. The red arrow indicates the convolution operation, the blue arrow indicates element-wise multiplication and the blue dashed arrow is the bottom-up pathway. The black dashed arrow between the output and input is the closed loop path. Recurrent and leaky connections are not shown to keep the network structure clear.}
\end{figure}
%%%%%%%%%%%%%%%%%%%%%%%%%%%%%%%%%%%%%%%%%%%%%%%%%%%%%%%%%%%%%%%%%%%%%%%%%%%%%%%%%%%
The model utilizes a convolutional neural network (CNN) and a deconvolutional neural network [9, 15]. However, unlike the CNN, every context layer in the P-MSTRNN model is composed of two different types of units that function differently from the units composing CNN layers. 

The first type is the feature unit and the other is the context unit. Both feature unit and context unit are based on the leaky integrator neural units which decay to the value of the previous time step, thereby enabling both types of units to reflect a temporal hierarchy that is not possible in a CNN [10]. The set of feature units in the same layer forms a feature map (FM) and context units form a context map (CM). FMs mainly contribute to spatial processing by receiving synaptic inputs from neighboring feature units. The CMs, on the other hand, contribute mainly to development of adequate dynamic information processing by taking advantage of their adjustable recurrent connectivity. The timescale that determines decay rate is differently set in all layers. The layers in the lower level have smaller time constants and the higher level layers have larger time constants. Therefore, the neural activity in the lower context layer is constrained to be faster while the one in the higher layers is constrained to be slower.

The forward dynamics of FMs for the generalized layer are shown in Equation (1) and (2). 
\begin{multline}
    \bm{\hat{f}}^{lp}_t = \Big(1-\frac{1}{\tau^{l}}\Big)\bm{\hat{f}}^{lp}_{t-1} + \frac{1}{\tau^l}\Big(\sum_{q=1}^{Q_{l+1}}\bm{f}^{(l+1)q}_{t-1} * \bm{k}^{lpq}_{ff} +\sum_{n=1}^{N_l}\bm{c}^{ln}_{t-1}*\bm{k}^{lpn}_{cf}
     + I*\bm{k}^{lp}_{if} + {b}^{lp}\Big)
\end{multline}

\begin{equation}
    \bm{f}^{lp}_t = 1.7159\tanh\Big(\frac{2}{3}\bm{\hat{f}}^{lp}_t\Big)
\end{equation}

where \(\hat{\bm{f}}^{lp}_t\) is the \(p^{th}\) FM’s internal state in the \(l^{th}\) layer at time step t, \(\hat{\bm{f}}^{lp}_{t-1}\) is the internal state of \(p^{th}\) FM in the \(l^{th}\) layer at t-1 time steps. \(\tau^l\) is the timescale in the \(l^{th}\) layer and \(Q_{l+1}\) is the number of FMs in the \({l+1}^{th}\) layer. The operation \(*\) is the convolution operator. \(\bm{k}^{lpq}_{ff}\) is the kernel connecting the \(q^{th}\) FM in the \({l+1}^{th}\) layer at t-1 time steps and \(p^{th}\) FM in the \(l^{th}\) layer at time step t. \(N_l\) is the total number of CM in the \(l^{th}\) layer and \(\bm{c}^{ln}_{t-1}\) is the activation of the \(n^{th}\) CM in the \(l^{th}\) layer at time step t-1. \(\bm{k}^{lpn}_{cf}\) is the kernel connecting the \(n^{th}\) CM in \(l^{th}\) layer at the previous time step t-1 and the \(p^{th}\) FM in the \(l^{th}\) layer at the current time step t. \(\bm{I}\) is the input data frame from outside of the network and \(\bm{k}^{lp}_{if}\) is the kernel between input and the current FM. \(b^{lp}\) is the bias of \(p^{th}\) FMs in the \(l^{th}\) layer. The first term of Equation (1) represents the decayed internal states of FMs from the previous time step t-1 with the decay rate \((1-\frac{1}{\tau^l})\). The second term is the input to the current FM from the activation value of FMs in the \({l+1}^{th}\) layer. The third term represents the input from the CMs in the same layer at the previous time step t-1. The fourth term indicates the data frame fed to the current FM. After the internal state of the current FM is computed, the activation value is obtained through a hyperbolic tangent activation function (Equation (2)). When the layer which the current \(p^{th}\) FM belongs to is other than the first layer, the term from the input data frame \(\bm{I}\) no longer exists. Also, highest layer internal dynamics do not include the second term in Equation 1 due to the absence of any higher layer to provide values. 

Equations (3) and (4) detail the forward dynamics of CMs. 
\begin{multline}
    \bm{\hat{c}}^{lm}_t = \Big(1-\frac{1}{\tau^{l}}\Big)\bm{\hat{c}}^{lm}_{t-1} + \frac{1}{\tau^{l}}\Big(\sum_{n=1}^{N_l}\bm{c}^{ln}_{t-1}\odot\bm{W}^{lmn}_{cc} +\sum_{q=1}^{Q_{l+1}}\bm{f}^{(l+1)q}_{t-1}\odot\bm{W}^{lmq}_{fc} \\ + 
    \sum_{r=1}^{R_{l-1}}\bm{f}^{(l-1)r}_{t-1}*\bm{k}^{lmr}_{fc}+{b}^{lm}\Big)
\end{multline}

\begin{equation}
    \bm{c}^{lm}_t = 1.7159 \tanh\Big(\frac{2}{3}\bm{\hat{c}}^{lm}_t\Big)
\end{equation}    

where \(\hat{\bm{c}}^{lm}_t\) is the internal state of the \(m^{th}\) CM in the \(l^{th}\) layer at time step t and \(\bm{c}^{ln}_{t-1}\) is the activation value of the \(n^{th}\) CM in the \(l^{th}\) layer at time step t-1. \(\bm{W}^{lmn}_{cc}\) is the recurrent weight connecting the \(m^{th}\) CM in time step t with the previous \(n^{th}\) CM in time step t-1.  \(\bm{W}^{lmq}_{fc}\) is the weight connecting the current \(m^{th}\) CM with the \(q^{th}\) FM in the \({l+1}^{th}\) layer at time step t-1. \(\bm{k}^{lmr}_{fc}\) is kernel connecting the \(r^{th}\) FM in \(l-1^{th}\) layer in t-1 step to current CM. \(\odot\) is the element-wise multiplication operator. The first term of Equation (3) is the leaky integrator input from the internal state of the CM from the previous time step t-1. The second term is the recurrent input to the current CM from the previous time step’s CM activation value. \(\bm{W}_{cc}^{lmn}\) is same size as CMs in the same layer. Accordingly, the activation values of the previous CM are multiplied with \(\bm{W}_{cc}\) in an element-wise manner. In the CMs, this second term represents the recurrency and reinforces the temporal processes of the network. The third term represents the input from the activation value of FMs in the \({l+1}^{th}\) layer. The fourth term is from the lower layer FMs. This term utilizes the bottom-up pathway allowing input data frames from outside of the network to be processed through all layers of the network. For the internal state calculation of the CMs, if the layer is the highest layer, then there is no input from the upper l+1 layer and the third term in Equation (3) is unused. Likewise, if the layer is the lowest layer, then an input from the lower layer does not exist and the fourth term in Equation (3) is unused. After the internal state is calculated, an activation value is obtained using the hyperbolic tangent function in Equation (4). 

When calculating the convolution, there are some cases in which the input map size is smaller than the size of the output map. In these cases, zero-padding is used for the input maps. As for the element-wise multiplication, the map and weight sizes are always the same. 

In the output layer, the output is calculated as,
\begin{equation}
    \bm{\hat{O}}_t = \sum_{q=1}^{Q_1}(\bm{f}^{1q}_t*\bm{k}^q_{fo} + {b}_o)
\end{equation}
    
\begin{equation}
    \bm{O}_t = 1.7159\tanh\Big(\frac{2}{3}\bm{\hat{O}}_{t}\Big)
\end{equation}
where \(\hat{\bm{O}}_t\) is the internal state of the output layer, \(Q_1\) is the number of FMs in the \(1^{st}\) layer. \(\bm{k}^q_{fo}\) is the kernel connecting the \(q^{th}\) FM in the \(1^{st}\) layer with the output map. \(b_o\) is the bias for the output map. The internal state of the output is calculated from the FMs in the first layer by convolution, and the activation value of the output layer is obtained by applying the scaled hyperbolic tangent function as suggested by LeCun et al. [16]. 

\subsection{Learning, generation and recognition}
The open loop generation method was used for training the network. When the network utilizes the open loop generation, the network receives the current input frame from the dataset and generates a (single or multiple) step prediction as an output frame for each step. We adopted Mean Square Error (MSE) for the cost function. A conventional back-propagation through time (BPTT) method was used for training the network. The weights, kernels, biases and initial states of the context layers were optimized using gradient descent. 
%After the prediction for the current input is obtained, the difference between the output frame and the prediction target frame at time step t is calculated, resulting in the error for all time steps. This process is shown in the Equation below. 

%\begin{equation}
%    E = \frac{1}{T}\sum_{t=1}^{T}E_t
%\end{equation}

%\begin{equation}
%    E_t = \frac{1}{XY}\sum_{i}^{X}\sum_{j}^{Y}(O^{*}_{ij}-O_{ij})^2
%\end{equation}

%where \(E_t\) is the average error per pixel at time step t and \(E\) is the average error per pixel for total time steps T. \(\bm{O}^*_{ij}\) is the target pixel value in the (i,j) position of the frame and \(\bm{O}_{ij}\) is the output pixel in the (i,j) position of the frame. The error calculated through this process is then used to optimize the parameters. 

%%%%%%%%%%%%%%%%%%%%%%%%%%%%%%%%%%%%% FIGURE x %%%%%%%%%%%%%%%%%%%%%%%%%%%%%%%%%%%%%%
%\begin{figure}[h]
%  \centering
%  \includegraphics[scale=0.315]{figure1AB_new}%[scale=0.3]
%  \caption{}
%\end{figure}
%%%%%%%%%%%%%%%%%%%%%%%%%%%%%%%%%%%%%%%%%%%%%%%%%%%%%%%%%%%%%%%%%%%%%%%%%%%%%%%%%%%

Along with open loop generation, closed loop generation was also used for training the network. The closed loop generation method is a scheme in which next step output prediction is computed by using a copy of the output prediction from the previous step as the current input. In the closed loop method, error is also calculated in comparison to the target signals. During training, error from open loop generation was used to optimize network parameters. As the training proceeded, both open loop error and closed loop error decreased. However, closed loop error was always higher than that of open loop because closed loop prediction generates accumulated error over time steps without correction from factors outside of the network. Therefore, network training was terminated when the closed loop error reached a predefined lower bound threshold. At this point, the network was guaranteed to successfully perform both open loop generation and closed loop generation. 

In order for the network to learn to generate multiple data sequences, it must infer optimal initial states of context units in all layers for each sequence, as well as optimal connectivity weights. Inferred initial states for each training sequence represent intentions to generate corresponding sequence. 

%%%%%%%%%%%%%%%%%%%%%%%%%%%%%%%%%%%%% FIGURE x %%%%%%%%%%%%%%%%%%%%%%%%%%%%%%%%%%%%%%
\begin{figure}[h]
  \centering
  \includegraphics[scale=0.34]{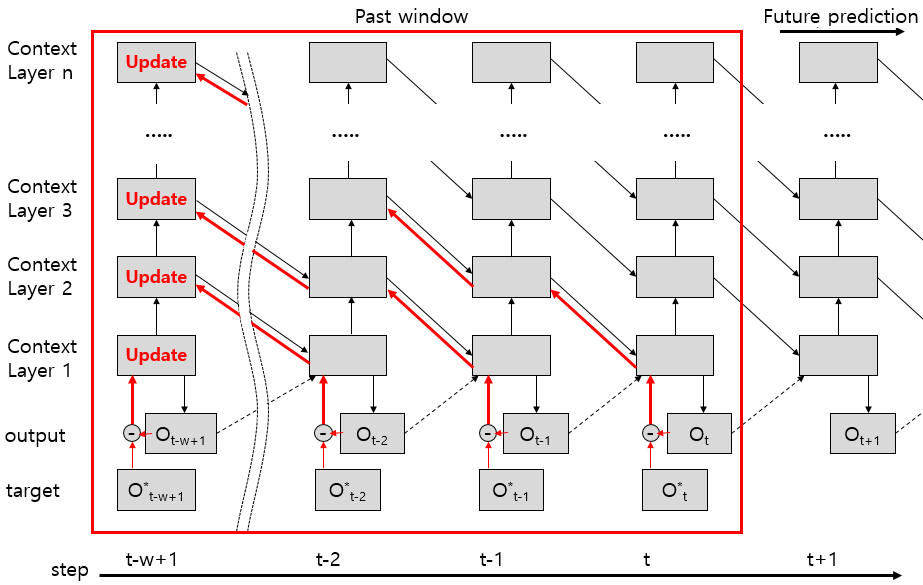}%[scale=0.37]
  \caption{Description of imitative synchronization via error regression. The black arrow indicates the forward process. The red arrow shows the prediction error flowing backward only within the window represented by red box. At current time step t, optimal prediction for next steps is calculated by adjusting initial states of the window. The output \(O_t\) is the prediction of the next step input and \(O^*_t\) is the target. }
\end{figure}
%%%%%%%%%%%%%%%%%%%%%%%%%%%%%%%%%%%%%%%%%%%%%%%%%%%%%%%%%%%%%%%%%%%%%%%%%%%%%%%%%%%

The novelty of this model is that the trained network can be used for both generation and recognition of sequence patterns. Once the network is trained, intended patterns are generated through the top-down pathway by setting the intention states of context units. The intention then propagates to the lower layers, eventually producing the intended or predicted image in a pixel level.
On the other hand, current input patterns can be recognized by inferring optimal intention states of context units. The process, imitative synchronization via error regression which will be explained later, is performed by first going through the bottom-up pathway for recognition to find the optimal internal states. Optimal prediction and generation are then achieved through the top-down pathway. 

In the framework of neuroscience, the brain receives sensory input by a bottom-up pathway and predicts the next input through a top-down path at every moment. When the actual input is different from what the brain predicted, the brain is surprised and modifies its prediction by reflecting the discrepancies. In this way, the brain always accurately predicts the future and actively interacts with the outer world [1, 14]. 
The idea of imitative synchronization via error regression is the same as the way the brain recognizes and predicts sensory inputs. 

%The future input must be predicted by adapting network's internal states for minimizing error between predictions and actual future inputs. 

Suppose the network is already trained to generate several types of human movement patterns in a pixel level video sequence given corresponding initial states. 
%The network now attempts to predict future so that it can actively imitate incoming streams of human movement patterns from an online camera at every moment. 
The network now attempts to actively imitate incoming streams of human movement patterns from an online camera at every moment by predicting future frames. 
This task is called imitative synchronization. 
To achieve imitative synchronization, the network must predict the next input pattern based on the recognized intention of the current input. The recognition process is performed by finding the network's optimal internal states that minimize error between predictions and actual inputs. This error minimizing method is called error regression.
During the test, if the human suddenly changes current movement pattern to another type of pattern, the network must quickly read the intention of the changed pattern and modify its generation in a pixel level. This active recognition and generation task of imitative synchronization can be achieved by error regression.
%The recognition here means to find intention state to the current sensory input. 

Through the error regression, for recognizing the input at the current time step \(t\), thereby producing optimal prediction for future steps, the network utilizes past information. 
%The network tries to produce output imitating input history that is already known to the network by adapting its internal states. 
The network adapts its internal states by trying to imitate input history that is already known to the network.
%produce output imitating input history that is already known to the network by adapting its internal states. 
When the network produces optimal reconstruction outputs for past inputs, future prediction is also made based on the same intention the network has. 
%tries to generate closed loop output that is closest to the actual input history. 
This process, error regression, occurs only within the temporal window from time step \(t-w+1\) to \(t\). 
The network first produces closed loop prediction starting from the time step \(t-w+1\) to \(t\) and error is calculated by comparing the actual input history and the output of the network in the same range of time. The error is then back propagated through time (BPTT) until it reaches the first step of window at \(t-w+1\), which is called intention states. 
%This process occurs only within the temporal window from time step \(t-w+1\) to \(t\) utilizing past information. 

%at the onset of immediately prior time steps by way of which the test sequence pattern can be reconstructed through closed-loop generation with minimal error in the temporal window.
%For this purpose, error BPTT is performed only within the immediate temporal window of \textit{w} steps. 
This scheme is shown in Figure 2 in detail. In the figure, the prediction error is back propagated through the path presented as red arrows and finally reaches initial states inside the temporal window at time step \(t-w+1\). The initial states are then modified to produce minimum reconstruction error between closed loop generation and the actual input history within the window while maintaining weights and biases unchanged. Until the prediction error decreases to produce an output pattern close to the input history, forward and backward propagation and modification of intention states are iterated multiple times inside the temporal window. 
%This is analogous to finding the network's intention that is closest to that of incoming current input. 
When the error is low enough that the network successfully imitates input history, prediction for future frame \(t+1\) is computed based on modified internal states. After that, the temporal window shifts one step forward in order to cover next time steps. 
In this manner, the test input pattern can be recognized in an on-line manner step by step and therefore allowing active prediction as close as possible to the input pattern. 
%This is the Imitative synchronization via the error regression of the internal states of the context units.

\section{Experiment}
An experimental dataset was designed to reveal the spatio-temporal structure of the proposed P-MSRTNN. The dataset consists of whole body human movement patterns generated by following a hierarchically defined movement syntax. Sub-primitives (arms and legs) were defined first. Each whole body movement primitive was composed of these sub-primitives, with sub-primitives being shared by all movements. 
Experiment 1 analyzed the self-organization of the spatio-temporal hierarchy as the model learned one subject's six primitive movements. After training these primitives, the network was trained with additional (previously unlearned) concatenations of prior-learned primitive patterns. Experiment 2 examined the recognition capability of the trained model. Specifically, how the robustness of recognition is dependent on variance in training exemplar patterns was tested by changing the number of subjects preparing training patterns. 
%%%%%%%%%%%%%%%%%%%%%%%%%%%%%%%%%%%%% FIGURE 2 3 %%%%%%%%%%%%%%%%%%%%%%%%%%%%%%%%%%%%%%
% \begin{figure}[!tbp]
%   \centering
%   \begin{minipage}[b]{0.45\textwidth}
%     \includegraphics[width=\textwidth]{figure2_new}
%     \caption{Sub-primitives of arms and legs.}
%   \end{minipage}
%   \hfill
%   \begin{minipage}[b]{0.45\textwidth}
%     \includegraphics[width=\textwidth]{figure3}
%     \caption{Examples of closed loop primitives generation. (P1, P3 and P5 from the top row)}
%   \end{minipage}
% \end{figure}

\begin{figure}[h]
  \centering
  \includegraphics[scale=0.3]{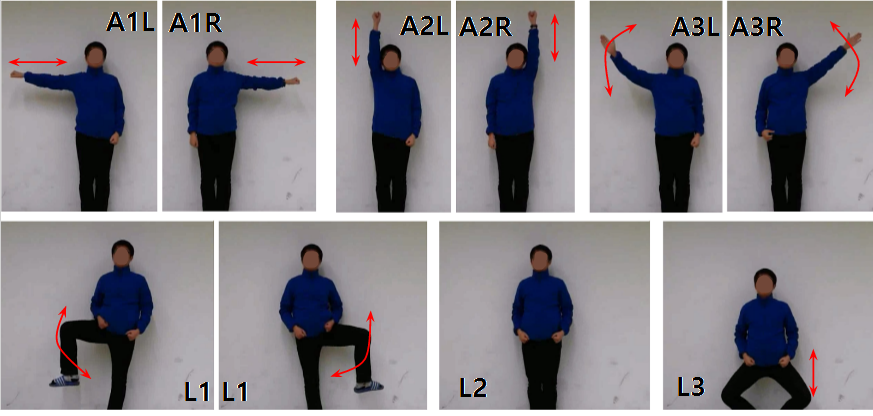}%0.35
  \caption{Sub-primitives of arms and legs.}
\end{figure}
  
\begin{figure}[h]
	\centering
    \includegraphics[scale=0.25]{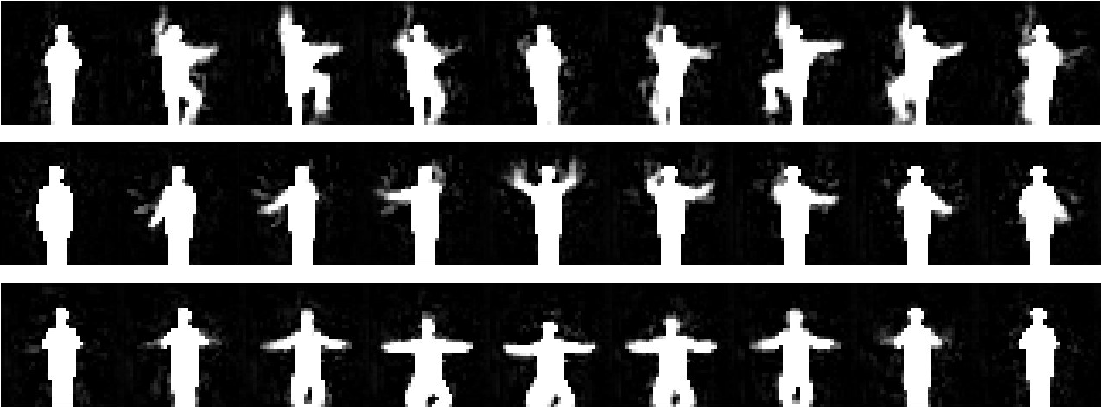}%0.3
    \caption{Examples of closed loop primitives generation. (P1, P3 and P5 from the top row)}
\end{figure}
%%%%%%%%%%%%%%%%%%%%%%%%%%%%%%%%%%%%%%%%%%%%%%%%%%%%%%%%%%%%%%%%%%%%%%%%%%%%%%%%%%%
\subsection{Dataset}
The dataset for this experiment consists of six whole body movement patterns. Each whole body movement pattern was hierarchically generated by combining predefined sub-primitives using legs and arms. Figure 3 describes the sub-primitives. There are three types of arm sub-primitives. Sub-primitive 1 (A1) is laterally extending arms. Sub-primitive 2 (A2) is vertically extending arms . Sub-primitive 3 (A3) is drawing a large circle with arms. In action space, these arm sub-primitives are represented as A1R (arm sub primitive 1, right), A1L(arm sub primitive 1, left), A2R, A2L, A3R and A3L. Leg sub-primitives appear in the second row of Figure 3. There are three types of leg sub-primitives. The first leg sub-primitive (L1) is raising the right and left leg alternatively. The second leg sub-primitive (L2) is standing still, moving neither leg. The third leg sub-primitive (L3) is bending both legs. All are shown in Figure 3. 
There are a total of six whole body action primitives. Their syntax is presented in Table 1, 
%%%%%%%%%%%%%%%%%%%%%%%%%%%%%%%%%%%%% TABLE 1 %%%%%%%%%%%%%%%%%%%%%%%%%%%%%%%%%%%%%%
\begin{table*}[]
\centering
\caption{Hierarchical syntax of action primitives}
\label{my-label}
\begin{tabular}{@{}ccccccccccccc@{}}
\toprule
                     & \multicolumn{2}{c}{P1}       & \multicolumn{2}{c}{P2}         & \multicolumn{2}{c}{P3}       & \multicolumn{2}{c}{P4}         & \multicolumn{2}{c}{P5}       & \multicolumn{2}{c}{P6}         \\ \midrule
\multirow{3}{*}{Arm} & Left         & Right         & Left          & Right          & Left         & Right         & Left          & Right          & Left         & Right         & Left          & Right          \\ \cmidrule(l){2-13} 
                     & A2L          & A1R           & A1L           & A2R            & A3L          & A3R           & A3L           & A3R            & A1L          & A1R           & A2L           & A2R            \\
                     & \multicolumn{2}{c}{Co-phase} & \multicolumn{2}{c}{Anti-phase} & \multicolumn{2}{c}{Co-phase} & \multicolumn{2}{c}{Anti-phase} & \multicolumn{2}{c}{Co-phase} & \multicolumn{2}{c}{Anti-phase} \\ \midrule
Leg                  & \multicolumn{2}{c}{L1}       & \multicolumn{2}{c}{L2}         & \multicolumn{2}{c}{L1}       & \multicolumn{2}{c}{L2}         & \multicolumn{2}{c}{L3}       & \multicolumn{2}{c}{L3}         \\ \bottomrule
\end{tabular}
\end{table*}
%%%%%%%%%%%%%%%%%%%%%%%%%%%%%%%%%%%%%%%%%%%%%%%%%%%%%%%%%%%%%%%%%%%%%%%%%%%%%%%%%%%
where the term "co-phase" involves two arm actions performed at once, and "anti-phase" involves arm actions performed alternatively. As can be seen in the table, every sub-primitive (arm and leg) was utilized two times during every whole-body movement. For training data, each whole body primitive was repeated for six cycles. The concatenated data for additional learning consists of P1 and P5 alternated three times (P1-P5-P1-P5-P1-P5) with each primitive repeated for three cycles. The length of primitive sequences is around 100 steps each and concatenated data are around 300 steps long each. The size of pixel is 36 by 36.

%%%%%%%%%%%%%%%%%%%%%%%%%%%%%%%%%%%%% FIGURE 4 %%%%%%%%%%%%%%%%%%%%%%%%%%%%%%%%%%%%%%
\begin{figure}[h]
  \centering
  \includegraphics[scale=0.2]{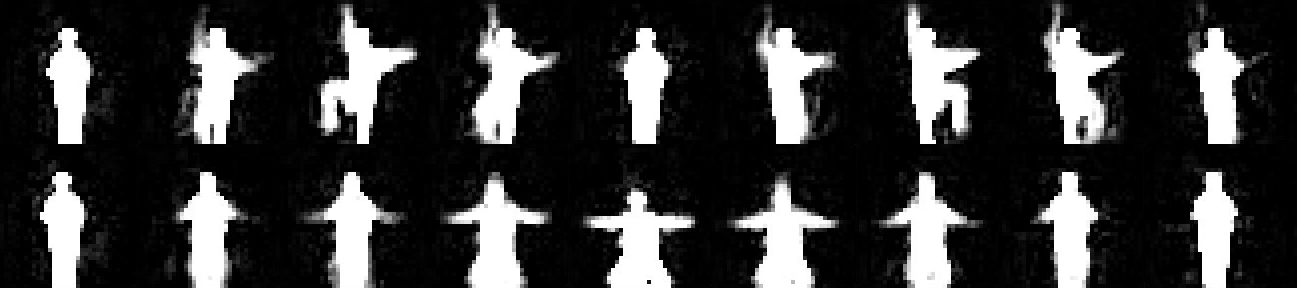}
  \caption{Example of closed loop combination sequence generation. The first row (P1) is followed by the second row (P5)}
\end{figure}
%%%%%%%%%%%%%%%%%%%%%%%%%%%%%%%%%%%%%%%%%%%%%%%%%%%%%%%%%%%%%%%%%%%%%%%%%%%%%%%%%%%

\subsection{Experiment 1}
In this experiment, characteristics of the proposed P-MSTRNN will be scrutinized. In particular, self-organized spatial and temporal hierarchy will be handled with a previously designed dataset. 
The table in the supplementary site (\nolinkurl{https://sites.google.com/site/academicpapersubmissions/p-mstrnn}) specifies the network parameters used for this experiment. The zeroth layer is the input/output layer and the remaining layers are context layers. All of the weights and the kernel sizes were defined according to the maps with which the weights or kernels connected. To train the network, the learning rate was set to 0.001.

In experiment 1, training proceeded in two stages. The network was first trained with six basic primitives. When the closed loop error decreased to the predefined level, the first training stage was terminated. Figure 4 shows an example of reconstructed images from closed loop output generation for three types of primitives. All the different types of primitive sequences, whose length is around 100 steps, were successfully generated without input from outside of the network. In the second stage of training, the trained network from stage 1 learned additional concatenations of the six primitives. As in stage 1, stage 2 was terminated when the closed loop error decreased to a certain level. Figure 5 represents a transition example of closed loop reconstruction of additional data from P1 to P5, and shows that the closed loop generation of additional concatenated data was successful for around 300 steps (See videos 1 and 2 in the supplementary material, \nolinkurl{https://sites.google.com/site/academicpapersubmissions/p-mstrnn}). The neural activities of this network for the six basic primitives and additional data are presented in Figure 6. 
The whole contexts' neural activities for all time steps were collected when the network generated each pattern. Dimensionality reduction was then performed on the collected data. Eventually, the neural activities are presented as two dimensional trajectories as in Figure 6. This way of representing neural activities is also applied to Figure 7 and Figure 9.
%%%%%%%%%%%%%%%%%%%%%%%%%%%%%%%%%%%%% FIGURE 5 %%%%%%%%%%%%%%%%%%%%%%%%%%%%%%%%%%%%%%
\begin{figure}[h]
  \centering
  \includegraphics[scale=0.450]{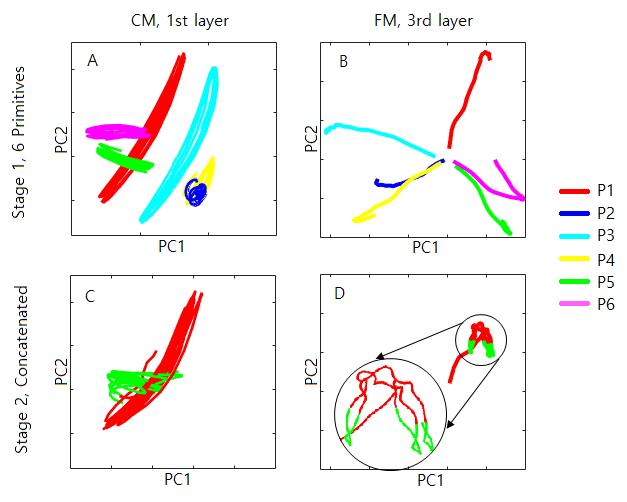}%0.55
%  (A)    \qquad \qquad \qquad  \qquad\qquad\qquad\qquad   (B)
  \caption{Neural activations of different layers. Different colors represent different whole body primitives (Red-P1, Blue-P2, Cyan-P3, Yellow-P4, Green-P5, Magenta-P6). Dimensionality reduction was performed by PCA. The x-axis is the first principal component, and the y-axis is the second principal component. (A)-neural activity of six primitives from CM in the first layer. (B)-neural activity of six primitives from FM in the third layer. (C)-neural activity of concatenated sequence from CM in the first layer. (D)-neural activity of concatenated sequence from FM in the third layer.}
\end{figure}
%%%%%%%%%%%%%%%%%%%%%%%%%%%%%%%%%%%%%%%%%%%%%%%%%%%%%%%%%%%%%%%%%%%%%%%%%%%%%%%%%%%

Figure 6.A plot shows neural activities of the CMs in the first layer corresponding to the six primitives and Figure 6.B shows neural activities in the third layer corresponding to the six basic primitives. Figure 6.C shows their neural activities corresponding to additional concatenated data in the first layer of CMs and Figure 6.D is with additional data from third layer FMs. The six cyclic patterns were developed in Figure 6.A while convergence toward six fixed points (starting from near the center and heading to the far region from the center) is observed in Figure 6.B. The resulting figures overall show the lower layers deal with more detailed information of the input pattern and the higher layers represent abstract or intentional information. 
In Figure 6.C, concatenated data activations of P1 and P5 appear in similar positions and shapes as the basic primitive shapes of P1 and P5 activations in Figure 6.A. Since the combination sequence was the alternation of two primitives, the trajectories of Figure 6.C also alternate from one to another. 
This implies that when the network already trained with simple primitives goes through additional learning using combination patterns, it reuses firstly learned primitives to organize concatenated primitives. This is why the neural activities of combination patterns presented in Figure 6.C (red and green) are similar to the activities of simple primitives in Figure 6.A  (red and green).
In Figure 6.D, activity converges to a cyclic pattern rather than to a fixed point. It was found that this cyclic neural activity in the 3rd layer induces cyclic neural activity corresponding to cyclic switching between P1 and P5 in the 1st layer, as shown in Figure 6.C. Similar to the analysis of lower layer (Figure 6.A and C), the higher layer's intention of concatenated pattern (Figure 6.D) also recycles the intention of simple primitives (Figure 6.B). 
We conclude that the model developed a temporal hierarchy by utilizing different time scales embedded in layers, enabling additional learning of combinatorial patterns to be achieved through the reuse of simple primitives.
%%%%%%%%%%%%%%%%%%%%%%%%%%%%%%%%%%%%% FIGURE 6 %%%%%%%%%%%%%%%%%%%%%%%%%%%%%%%%%%%%%%
\begin{figure}[h]
  \centering
  \includegraphics[scale=0.450]{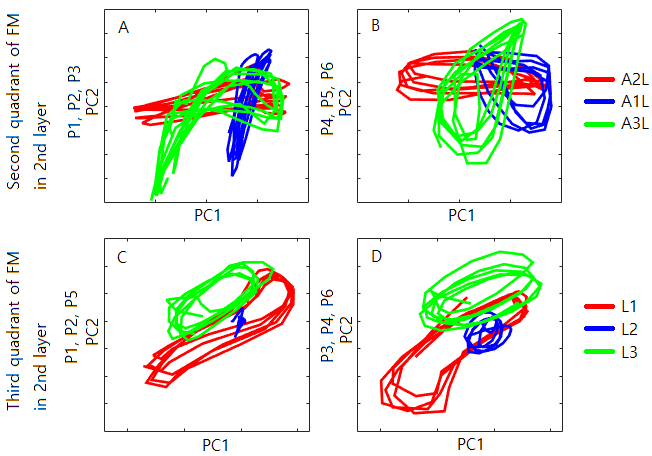}\\
%  (A)    \qquad \qquad \qquad\qquad\qquad\qquad\qquad   (B)
  \caption{Neural activations of different quadrants in the same map. Different colors represent the different sub-primitives. The plot data were obtained from the FM in the second layer and PCA was performed. (A)-Second quadrant of the FM (Red-P1, Blue-P2, Green-P3). (B)-Second quadrant of the FM (Red-P6, Blue-P5, Green-P4). (C)-Third quadrant of the FM (Red-P1, Blue-P2, Green-P5). (D)-Third quadrant of the FM (Red-P3, Blue-P4, Green-P6).}
\end{figure}
%%%%%%%%%%%%%%%%%%%%%%%%%%%%%%%%%%%%%%%%%%%%%%%%%%%%%%%%%%%%%%%%%%%%%%%%%%%%%%%%%%%

To uncover the spatial hierarchy developed in the model, we conducted another analysis with the same trained network. FMs were divided into four quadrants and neural activities were recorded separately. Figure 7 shows one example of the resulting plots from the FM in the second layer. Figure 7.A,B present the neuronal activity of the second quadrant of the map which is the upper left part of the map whose receptive fields correspond to left arm activation. Figure 7.A plots red trajectory for the data from P1 and Figure 7.B represents P6 data in red. Looking at Table 1, P1 and P6 share the same left arm sub-primitive (A2L) and thus the corresponding neural activations are plotted in the same positions and in the same shapes. A similar analysis also applies to P2 and P5. In the same figures, P2 is presented in blue in Figure 7.A and P5 is drawn in blue in Figure 7.B. According to the syntax comprising primitives, P2 and P5 share the same left arm movement (A1L). Also, for P3 and P4 trajectories drawn in green, the same phenomenon is observed. In the case of Figure 7.C,D, data are obtained from the third quadrant whose receptive fields correspond to left leg activation. Action primitives P1 (red in the Figure 7.C) and P3 (red in the Figure 7.D) share leg sub-primitive L1. The pair, P2 and P4 (blue), and the pair P5 and P6 (green) also share leg sub-primitives. It is clearly shown that these pairs exhibit similar shapes of activation trajectories located in similar positions of the phase plots. This analysis suggests that when whole body primitives share sub-primitives, neural activation in the same quadrants is the same, implying that P-MSTRNN developed a spatial hierarchy combining sub-primitive limb-specific movement patterns.

\subsection{Experiment 2}
In experiment 2, we tested the model's recognition ability. As explained previously, a network's recognition performance is related to the ability to find the closest initial states (intention) corresponding to the current input, thereby actively generating incoming input. Therefore, as the network produces a more accurate prediction output, the network is regarded as having the better recognition ability. With the scheme of imitative synchronization via error regression, P-MSTRNN performs not only generation but also robust recognition in one model. 

In this experiment, the network structure and parameter settings were the same as those in experiment 1. However, the error regression adaptation rate, which is the same as the learning rate in the temporal window was set to 0.1 and the window size was 30. The dataset used in this experiment consists of three types of whole body movements (P1, P4 and P5) performed by multiple subjects. The dataset exhibited roughly 15\% variance in movement speed amongst subjects who also differed in shape and height. 

The network was trained with training data and after that, imitative synchronization by error regression was conducted for test sequences by test subjects (see video3 in the supplementary material). 
For testing, subjects not included in training data performed a combination sequences of three basic primitives containing all of the possible transitions from one primitive to another (for example, P1-P4-P5-P1-P5-P4-P1). 

Figure 8 shows a representative snapshot of activation values during a transition from P1 to P5 in the five subjects training case.
The example transition between primitives occurs at around 660 time steps in Figure 8.A. Error rises sharply leading up to the transition, inducing modulation of context activity through error regression, finally resulting in the correct output pattern and transition. As can be seen in the first and the second rows of Figure 8.A, when the input pattern suddenly changes, the output fails to follow the input at first but eventually the network adapts  to produce a desirable output. After the transition, error falls close to zero. As illustrated in Figure 8.B, neural activations in the first layer show rapid changes compared to those of the fourth layer. This is due to the fact that time constants increase as the layer level increases. The first layer has the smallest time constant and shows relatively fast activation changes. In this way, the network achieves temporal hierarchy. 

We compared the recognition ability of the P-MSTRNN using error regression to other models. Other models were performed in a sensory entrainment manner. The sensory entrainment is a scheme where the trained network receives the input from the test dataset for each step and produces a corresponding prediction output. For comparison, we used the LSTM Future Predictor Model conditioned on the input proposed by Srivastava et al. (2016) [7]. Also ConvLSTM by Shi et al. (2015) [12] was compared. In addition to the two models, P-MSTRNN in the entrainment scheme was also compared to the error regression.

\begin{table}[]
\centering
\caption{Comparison for averaged MSE for different models.}
\label{my-label}
\begin{tabular}{llc}
\hline
\textbf{Type}                                                       & \textbf{Model}                & \textbf{\begin{tabular}[c]{@{}c@{}}MSE\\ (Multi-Subject Dataset)\end{tabular}} \\ \hline
\begin{tabular}[c]{@{}l@{}}Error\\ Regression\end{tabular} & P-MSTRNN                      & 0.0391                                                                        \\ \hline
                                                                    & P-MSTRNN                      & 0.0608                                                                        \\
Entrainment                                                         & LSTM Future Predictor {[}7{]} & 0.0490                                                                        \\
                                                                    & ConvLSTM {[}12{]}             & 0.0454                                                                        \\ \hline
\end{tabular}
\end{table}
%%%%%%%%%%%%%%%%%%%%%%%%%%%%%%%%%%%%%%%%%%%%%%%%%%%%%%%%%%%%%%%%%%%%%%%%%%%%%%%%%%%

In Table 2, the average Mean Square Error (MSE) is shown. As can be seen in the results, P-MSTRNN with the error regression showed the best recognition performance and yielded the most accurate prediction outputs. Part of the reason is that the test dataset contained complex transitions between primitives similar to that shown in Figure 5. All the models were trained with simple primitives performed by multiple subjects and tested on the combinations of those primitives. 

Most of the previous generative models dealing with video datasets predicted future frames by passively conditioning on all the past information.
For this reason, even though the previous models received input from the test sequence every step in an entrainment scheme, it was difficult to quickly adapt its internal states to sudden transitions of primitives in the test sequences. 
Therefore most of the error occurred when the primitives in the test sequence changed to one another. Also, the entrainment type of P-MSTRNN showed the worst results. This is because the other models using LSTM utilized forget gates and thus adapting their internal states was faster than conventional RNN units. Additionally, it took more time for the entrainment type P-MSTRNN to update its higher intention states. 
%Therefore only using the entrainment is not sufficient for recognizing the current input for all models. 

However, the error regression repeatedly and actively updates network's internal states to reflect a sudden transition and this is the reason for the robust recognition performance even though the model uses conventional RNN units. 
%We conclude here that although P-MSTRNN is only trained on basic primitives, it is possible to recognize sequential combinations of those primitives very well. 
We conclude here that only using the entrainment is not sufficient for recognizing the current input, but in the proposed model, although P-MSTRNN is only trained on basic primitives, it is possible to recognize sequential combinations of those primitives very well.

%%%%%%%%%%%%%%%%%%%%%%%%%%%%%%%%%%%%% figure7 %%%%%%%%%%%%%%%%%%%%%%%%%%%%%%%%%%%%%%
\begin{figure}[h]
  \centering
  \includegraphics[scale=0.5]{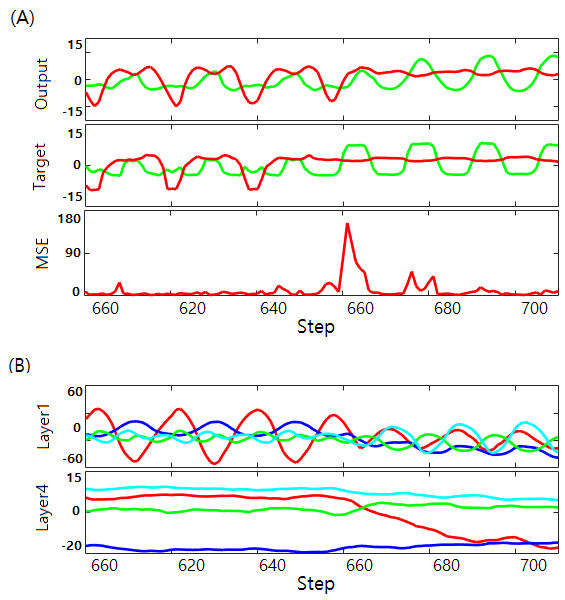}
  \caption{Results of imitative synchronization. The activations are the result of PCA. (A)-shows output, target and MSE. For the first and the second row, first principal component (PC) and second PC are shown in different colors. (B)-shows the neural activities of FM in layer 1 and FM in layer 4. Four different colors represent from the first PC to fourth PC. }
\end{figure}
%%%%%%%%%%%%%%%%%%%%%%%%%%%%%%%%%%%%%%%%%%%%%%%%%%%%%%%%%%%%%%%%%%%%%%%%%%%%%%%%%%%

In order to see the effect of variances in exemplar training data towards the network's recognition ability by error regression, we compared two cases. The first P-MSTRNN network was trained with a multi-subject dataset performed by multiple people and the second P-MSTRNN network was trained with only one subject's data. 
The error regressions were conducted for two networks and the results are shown in Table 3. 
%The average MSE per pixel in the two conditions (multi-subject train / single-subject train) according to the three test subjects were (0.0685 / 0.0941), (0.1019 / 0.1378) and (0.0813 / 0.0992). 
%%%%%%%%%%%%%%%%%%%%%%%%%%%%%%%%%%%%% table %%%%%%%%%%%%%%%%%%%%%%%%%%%%%%%%%%%%%%
\begin{table}[]
\centering
\caption{Comparison for averaged MSE for different dataset.}
\label{my-label}
\begin{tabular}{ccc}
\hline
\textbf{Model}                                                        & \textbf{\begin{tabular}[c]{@{}c@{}}MSE \\ (Multi-Subject Dataset)\end{tabular}} & \textbf{\begin{tabular}[c]{@{}c@{}}MSE \\ (Single-Subject Dataset)\end{tabular}} \\ \hline
\begin{tabular}[c]{@{}c@{}}P-MSTRNN\\ (Error \\Regression)\end{tabular} & 0.0391                                                                          & 0.0522                                                                           \\ \hline
\end{tabular}
\end{table}%%%%%%%%%%%%%%%%%%%%%%%%%%%%%%%%%%%%%%%%%%%%%%%%%%%%%%%%%%%%%%%%%%%%%%%%%%%%%%%%%%%

Table 3 shows the network trained under the multi-subject training condition yielded better results in imitative synchronization than the network trained under the single-subject training condition. This suggests that increased variance in training results in more robust recognition ability. This result is analogous to that of [11] which used lower dimensional data compared to the present study.

Under the multi-subject condition, identical movement primitives differed slightly from one another, and this variance was well preserved in the closed-loop output generation of dynamic images. Figure 9 shows the corresponding analysis of neural activity in different layers in the network for five subjects. (A) shows neural activations of CMs in the first layer and (B) shows those of CMs in the fourth layer. In the lower layer (Figure 9.A), the same primitives formed clusters. At the same time, variations of neural activation patterns corresponding to the different subjects in the same cluster are well preserved (note the different shapes and the positions of the trajectories). 
As in the analysis of experiment 1, higher layer intentions represented in terms of different fixed points in (B) determine the corresponding cyclic activation patterns in the lower level shown in (A) by means of parameter bifurcation. In training with multiple subjects as shown in Figure 9.B, three clusters emerge (one for each primitive), and fixed points vary within these clusters (reflecting subject variations). This implies that the higher level influences the lower level through the differentiation of fixed points within clusters of more or less robustly established primitives.
%%%%%%%%%%%%%%%%%%%%%%%%%%%%%%%%%%%%% FIGURE 8 %%%%%%%%%%%%%%%%%%%%%%%%%%%%%%%%%%%%%%
\begin{figure}[h]
  \centering
  \includegraphics[scale=0.5]{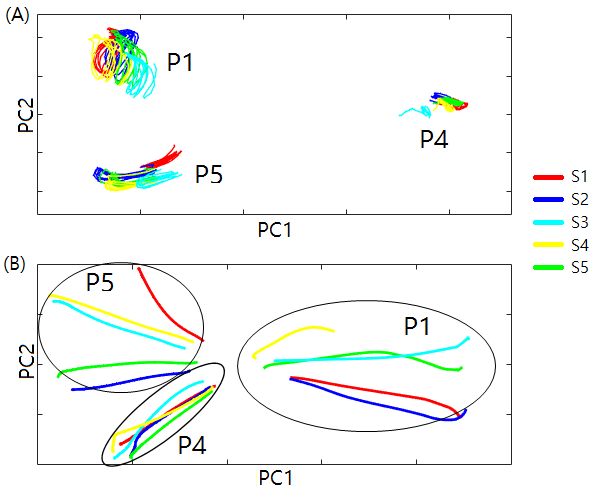}%0.5
%  (A)    \qquad \qquad \qquad  \qquad\qquad\qquad\qquad\qquad\qquad   (B)
  \caption{Neural activities of different layers. Different colors indicate different subjects. Three primitives with five subjects are plotted in both A and B. (A)-neural activations of CM of the first layer. (B)-neural activations of FM of the fourth layer. Dimensionality reduction is performed by PCA.}
\end{figure}

\section{Conclusion}
The current paper introduced a novel dynamic neural network model for both generating and recognizing dynamic visual image patterns at the pixel level within a predictive coding framework. P-MSRTNN is distinguished from other generative models in that generation and recognition are possible in one model. Our simulation experiments showed that the proposed model characterized by its multiple spatio-temporal scales property can learn to generate dynamic visual images representing human movement patterns through the development of an internal spatio-temporal hierarchy. The model can also robustly recognize test movement patterns containing multiple transitions performed by unfamiliar subjects through imitative synchronization by inferring intention states with the error regression scheme applied to context units.

Future studies should focus on scaling the model in terms of size of frames, as well as in terms of number and complexity of movement patterns. Also, applying the proposed model to the robotics domain is considered. Since the model utilizes temporal hierarchy, top-down and bottom-up pathway which are also used in the human brain, this model would be beneficial to robots to act more natural and human-like way. Moreover, robots with error regression scheme, which can be regarded as memory retrieving process, would be interesting.

% conference papers do not normally have an appendix

% use section* for acknowledgment
\section*{Acknowledgment}
This work was supported by a National Research Foundation of Korea (NRF) grant funded by the Korea government (MSIP) (No.2014R1A2A2A01005491). We thank Ahmadreza Ahmadi for his help with the CUDA program.

% trigger a \newpage just before the given reference
% number - used to balance the columns on the last page
% adjust value as needed - may need to be readjusted if
% the document is modified later
%\IEEEtriggeratref{8}
% The "triggered" command can be changed if desired:
%\IEEEtriggercmd{\enlargethispage{-5in}}

% references section

% can use a bibliography generated by BibTeX as a .bbl file
% BibTeX documentation can be easily obtained at:
% http://mirror.ctan.org/biblio/bibtex/contrib/doc/
% The IEEEtran BibTeX style support page is at:
% http://www.michaelshell.org/tex/ieeetran/bibtex/
%\bibliographystyle{IEEEtran}
% argument is your BibTeX string definitions and bibliography database(s)
%\bibliography{IEEEabrv,../bib/paper}
%
% <OR> manually copy in the resultant .bbl file
% set second argument of \begin to the number of references
% (used to reserve space for the reference number labels box)
\section*{References}
[1] Rao, R.P. and Ballard, D.H., 1999. Predictive coding in the visual cortex: a functional interpretation of some extra-classical receptive-field effects. {\it Nature neuroscience}, \textbf{2}(1), pp.79-87.

[2] Tani, J. and Nolfi, S., 1999. Learning to perceive the world as articulated: an approach for hierarchical learning in sensory-motor systems. {\it Neural Networks}, \textbf{12}(7), pp.1131-1141.

[3] Friston, K., 2005. A theory of cortical responses. {\it Philosophical Transactions of the Royal Society of London B: Biological Sciences}, \textbf{360}(1456), pp.815-836.

[4] Yamashita, Y. and Tani, J., 2008. Emergence of functional hierarchy in a multiple timescale neural network model: a humanoid robot experiment. {\it PLoS Comput Biol}, \textbf{4}(11), p.e1000220.

[5] Radford, A., Metz, L. and Chintala, S., 2015. Unsupervised Representation Learning with Deep Convolutional Generative Adversarial Networks. {\it arXiv preprint} arXiv:1511.06434.

[6] Lotter, W., Kreiman, G. and Cox, D., 2015. Unsupervised Learning of Visual Structure using Predictive Generative Networks. {\it arXiv preprint} arXiv:1511.06380.

[7] Srivastava, N., Mansimov, E. and Salakhutdinov, R., 2015. Unsupervised learning of video representations using LSTMs. {\it arXiv preprint} arXiv:1502.04681.

[8] Lee, H., Jung, M. and Tani, J., 2016. Characteristics of Visual Categorization of Long-Concatenated and Object-Directed Human Actions by a Multiple Spatio-Temporal Scales Recurrent Neural Network Model. {\it arXiv preprint} arXiv:1602.01921.

[9] LeCun, Y., Bottou, L., Bengio, Y. and Haffner, P., 1998. Gradient-based learning applied to document recognition. {\it Proceedings of the IEEE}, \textbf{86}(11), pp.2278-2324.

[10] Jung, M., Hwang, J. and Tani, J., 2015. Self-Organization of Spatio-Temporal Hierarchy via Learning of Dynamic Visual Image Patterns on Action Sequences. {\it PloS one}, \textbf{10}(7), p.e0131214.

[11] Ahamdi, A. and Tani, J., 2016, October. Towards Robustness to Fluctuated Perceptual Patterns by a Deterministic Predictive Coding Model in a Task of Imitative Synchronization with Human Movement Patterns. In International Conference on Neural Information Processing (pp. 393-402). Springer International Publishing.

[12] Xingjian, S.H.I., Chen, Z., Wang, H., Yeung, D.Y., Wong, W.K. and Woo, W.C., 2015. Convolutional LSTM network: A machine learning approach for precipitation nowcasting. In Advances in Neural Information Processing Systems (pp. 802-810).

[13] Lotter, W., Kreiman, G. and Cox, D., 2016. Deep Predictive Coding Networks for Video Prediction and Unsupervised Learning. arXiv preprint arXiv:1605.08104.

[14] Friston, K., Mattout, J. and Kilner, J., 2011. Action understanding and active inference. Biological cybernetics, 104(1-2), pp.137-160.

[15] Zeiler, M.D., Taylor, G.W. and Fergus, R., 2011, November. Adaptive deconvolutional networks for mid and high level feature learning. In 2011 International Conference on Computer Vision (pp. 2018-2025). IEEE.

[16] LeCun, Y.A., Bottou, L., Orr, G.B. and Müller, K.R., 2012. Efficient backprop. In Neural networks: Tricks of the trade (pp. 9-48). Springer Berlin Heidelberg.

% that's all folks
\end{document}